\documentclass[journal,twoside,web]{ieeecolor}
\usepackage{tmi}
\usepackage{cite}
\usepackage{amsmath,amssymb,amsfonts}
\usepackage{algorithmic}
\usepackage{graphicx}
\usepackage{textcomp}
\usepackage{algorithmic}
\usepackage{multirow}
\usepackage{multicol}
\usepackage{textcomp}
\usepackage[ruled,vlined]{algorithm2e}
\usepackage[usenames,dvipsnames]{xcolor}

\def\BibTeX{{\rm B\kern-.05em{\sc i\kern-.025em b}\kern-.08em
    T\kern-.1667em\lower.7ex\hbox{E}\kern-.125emX}}
\definecolor{mygreen}{rgb}{0, 0.69, 0.314}
\definecolor{myred}{rgb}{0.753, 0, 0}
\definecolor{myblue}{rgb}{0.004, 0.69, 0.941}

\author{Kanggeun~Lee
        and~Won-Ki~Jeong$^*$\thanks{*Corresponding author.}

\thanks{This article has been accepted for publication in IEEE Transactions on Medical Imaging. This is the author's version which has not been fully edited and content may change prior to final publication. Citation information: DOI 10.1109/TMI.2021.3096142.}
\thanks{Kanggeun Lee is with the school of Electrical and Conputer Engineering, UNIST, South Korea (e-mail: leekanggeun@unist.ac.kr).}
\thanks{Won-Ki Jeong is with the Department of Computer Science and Engineering, Korea University, South Korea (e-mail: wkjeong@korea.ac.kr).}}

\begin{document}
\title{ISCL: Interdependent Self-Cooperative Learning for Unpaired Image Denoising}

\maketitle  

\begin{abstract}
With the advent of advances in self-supervised learning, paired clean-noisy data are no longer required in deep learning-based image denoising. However, existing blind denoising methods still require the assumption with regard to noise characteristics, such as zero-mean noise distribution and pixel-wise noise-signal independence; this hinders wide adaptation of the method in the medical domain. On the other hand, unpaired learning can overcome limitations related to the assumption on noise characteristics, which makes it more feasible for collecting the training data in real-world scenarios. 
In this paper, we propose a novel image denoising scheme, Interdependent Self-Cooperative Learning (ISCL), that leverages unpaired learning by combining cyclic adversarial learning with self-supervised residual learning. Unlike the existing unpaired image denoising methods relying on matching data distributions in different domains, the two architectures in ISCL, designed for different tasks, complement each other and boost the learning process. To assess the performance of the proposed method, we conducted extensive experiments in various biomedical image degradation scenarios, such as noise caused by physical characteristics of electron microscopy (EM) devices (film and charging noise), and structural noise found in low-dose computer tomography (CT). We demonstrate that the image quality of our method is superior to conventional and current state-of-the-art deep learning-based unpaired image denoising methods. 
\end{abstract}


\begin{figure}[t]
\includegraphics[width=9cm,keepaspectratio]{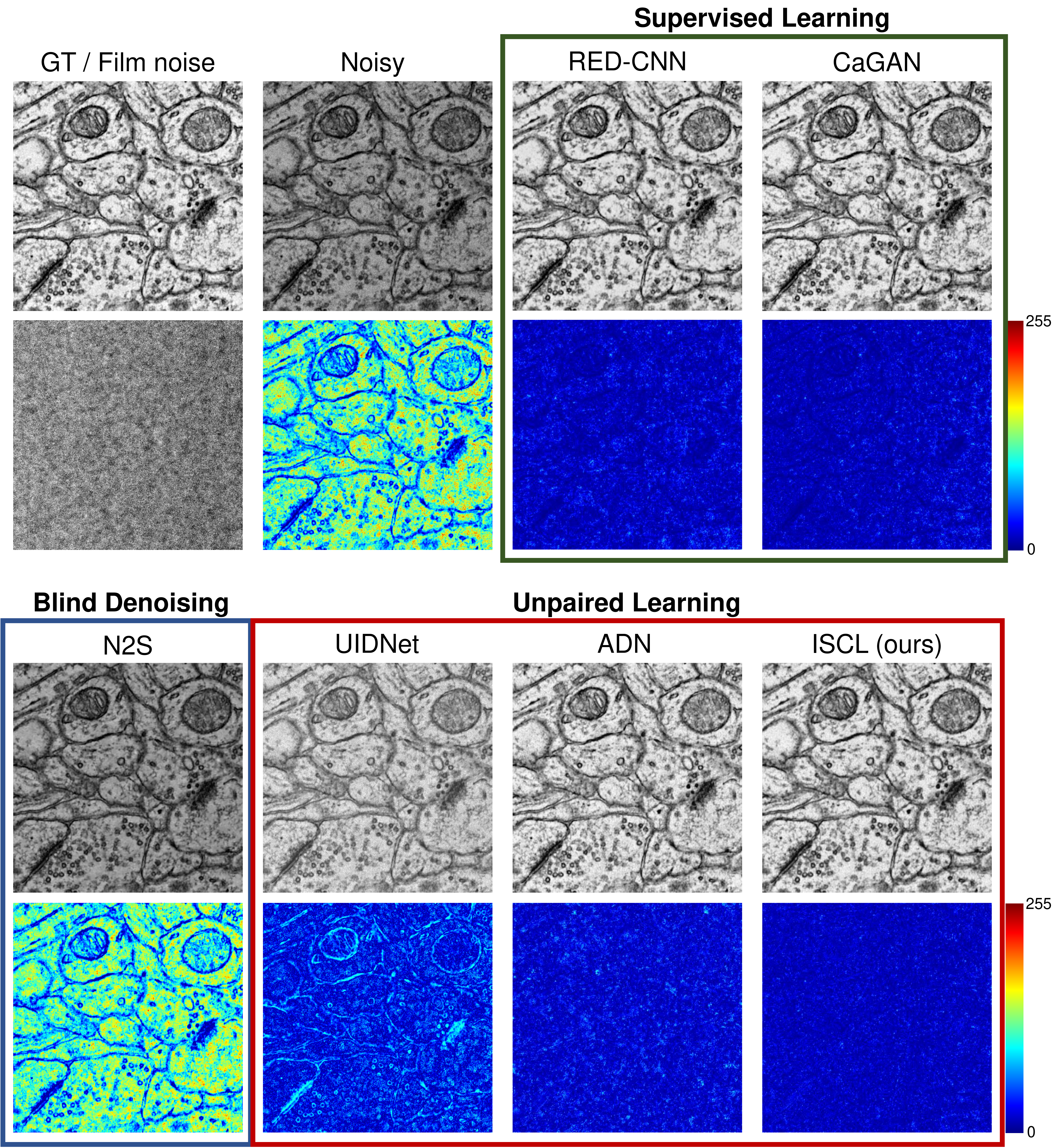}
\caption{An example of film noise removal in electron microscopy images. Top left: clean ground-truth image (top) and real film noise image (bottom). N2S\cite{noise2self} is a blind denoising method. RED-CNN~\cite{redcnn} and CaGAN~\cite{cagan} are supervised learning based denoising methods. UIDNet~\cite{hong2020end}, ADN~\cite{adn}, and our method (ISCL) are unpaired learning based approaches.}
   \label{fig:fig1}
\end{figure}

\section{INTRODUCTION}
\label{sec:introduction}
\IEEEPARstart{D}{enoising} is the low-level signal processing technique used to remove specific noise from noisy observation in order to improve the quality of signal analysis. 
Before deep learning gained its popularity, 
most image denoising research focused on leveraging image prior information, such as through non-local self-similarity~\cite{buades2005non,dabov2007,mairal2009non}, sparsity feature~\cite{bao2013fast, bao2015dictionary, elad2006image, papyan2017convolutional}, and total variation~\cite{vogel1996iterative,vese2004image,getreuer2012rudin}. 
In recent years, supervised learning methods using deep convolutional neural networks (CNNs) have surpassed the performance of prior-based denoising methods~\cite{zhang2017beyond, lefkimmiatis2018universal}. 
CNN models can learn to restore a clean target via paired training data without prior knowledge of image or noise. 
However, their performance is demonstrated only on well-known noise models. 
The main reason for this is that the training data (clean and noisy image pairs) are generated by adding noise for a given distribution to clean images.
Therefore, unconventional image degradation cannot be easily modeled, which makes the application of supervised learning difficult.

Recently, several self-supervised blind denoising methods~\cite{noise2self, noise2void, quan2020self2self} have shown promising results without the noise prior and the clean-noisy paired training data. 
The blind spot based approaches predict the clean pixel from the neighbor information of the target pixel based on the self-supervision training. 
However, these blind denoising methods require zero-mean noise distribution to apply the self-supervision loss. 
We observed that the state-of-the-art blind denoising and prior-based denoising methods tend to introduce incorrect brightness shifting 
for non-zero mean noise cases, as shown in Fig.~\ref{fig:fig1} (see BM3D and N2S results are still darker than the ground truth). 
In addition, noise should be pixel-wise independent under the given noisy observation to employ a blind spot scheme; this is not satisfied in unconventional noise observed in biomedical images.
For example, recent high-throughput automatic imaging using transmitted electron microscopy (TEM)~\cite{hildebrand_gridtape_2017, graham_gridtape_2018,Graham657346} uses electron-lucent support films, which introduce spatially inhomogeneous noise (i.e., film noise). 
In addition, prolonged exposure of electron beams onto the thin tissue section causes blob-like damage (i.e., charging noise) in scanning electron microscopy (SEM) images. 
See the leftmost images in the second and fourth rows in Fig.~\ref{fig:fig5} for each noise example.
Such imaging artifacts do not satisfy the necessary conditions for blind denoising.

The primary motivation behind our proposed work stems from the recent advances in unpaired image denoising~\cite{minh2019removing, hong2020end}. 
Quan \textit{et al.}~\cite{minh2019removing} demonstrated superior denoising performance on electron microscopy (EM) images without paired training data by leveraging three-way cyclic constraints with adversarial training. 
However, this method requires real noise pattern images (e.g., taking an empty film image, etc.), which is not always feasible in a real scenario (such as low-dose CT (LDCT)).
More recently, UIDNet~\cite{hong2020end} proposed an end-to-end denoising network trained by clean-pseudo noisy pair images where pseudo noisy images are automatically generated via a generative model. %
However, they only used a simple (weak) generative model to learn the noise distribution from examples, which is insufficient for unconventional noise, as in EM images (see Fig.~\ref{fig:fig1}). 
Our proposed method addresses the above problems via \textit{cooperative learning} -- multiple inter-domain mapping functions are trained together in a cooperative manner, which serves as stronger constraints in unsupervised training. 
In this paper, we propose a novel image denoising framework, Interdependent Self-Cooperative Learning (ISCL), to restore the clean target from the noise-corrupted image without using either paired supervision or prior knowledge of noise distribution. 
ISCL consists of two components, CycleGAN~\cite{zhu2017unpaired}-based denoiser learning, and pseudo-label based residual learning of a noise extractor, to boost the performance self-interdependently via cooperative learning. 
For training the denoiser with the proposed constraints, the noise extractor will assist the learning of the denoiser under the proposed loss. 
Conversely, the noise extractor will be trained by pairs of pseudo-clean and noisy with the noise consistency.
The main contributions of our work can be summarized as follows: 
\begin{enumerate}
\item We propose ISCL, an unpaired image denoiser based on a novel mutually adaptive training that integrates two different tasks. ISCL shows better denoising performance with faster convergence.
\item The proposed novel loss functions (i.e., bypass-consistency, discriminator boosting, and noise-consistency) promote the convergence toward an ideal denoiser.
\item We demonstrate that the proposed architecture of ISCL is optimal with respect to its model size; it can achieve superior performance using only a fraction of network parameters compared to state-of-the-art unpaired image denoising methods.

\end{enumerate}

\begin{figure*}[t]
\includegraphics[width=18cm,keepaspectratio]{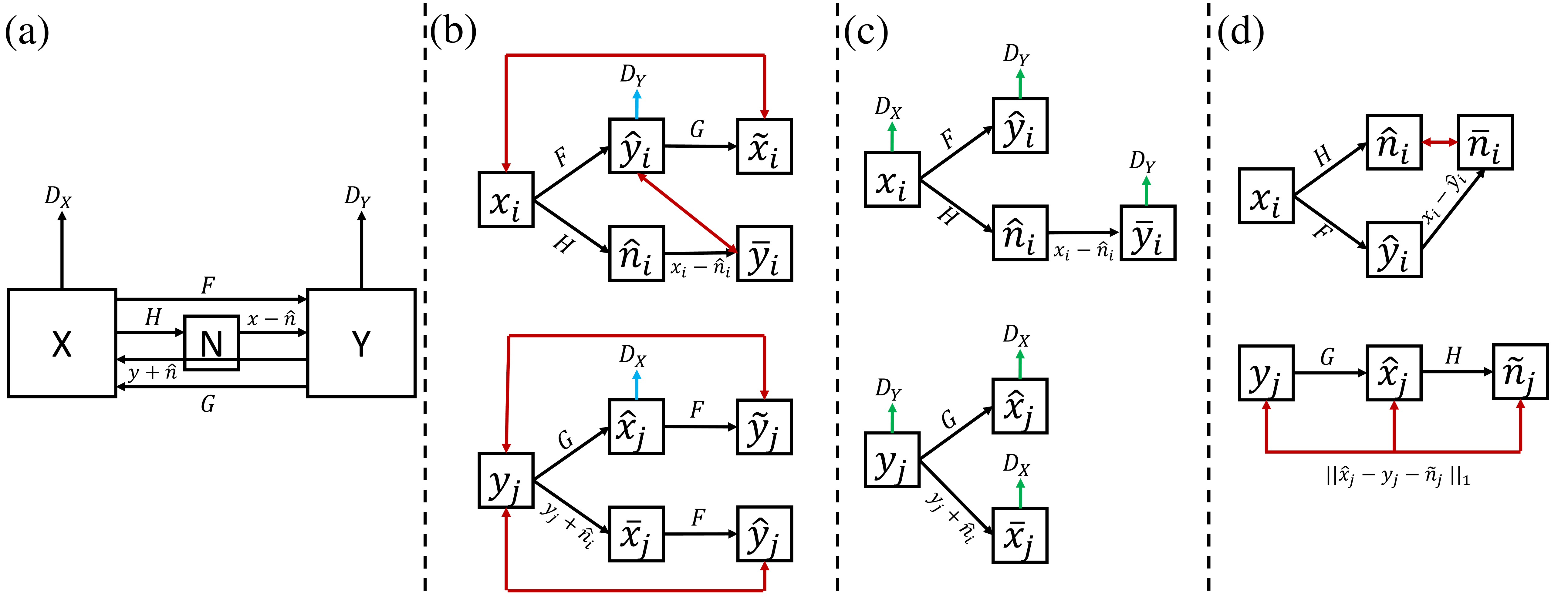}
\caption{(a) Flow graph of ISCL. Our proposed scheme has three mapping functions, $F: X\rightarrow Y$, $G: Y\rightarrow X$, and $H: X\rightarrow N$ with two discriminators, $D_X$ and $D_Y$.
$X$, $Y$, and $N$ are domains of noisy target, clean source, and noise of noisy target, respectively. (b) Training process of two mapping functions, $F$ and $G$. The \textcolor{myblue}{blue} means that the $F$ and $G$ are encouraged by outputs of $D_X$ and $D_Y$ for translation from one domain to the other as in adversarial learning. The \textcolor{myred}{red}  indicates the mean absolute error (MAE) between two instances as a cycle consistency. (c) Training process of two discriminators, $D_X$ and $D_Y$. Each discriminator can learn to distinguish between real and fake generated by $F$, $G$ and $H$. 
The \textcolor{mygreen}{green} indicates the inputs for the learning of each discriminator. (d) Training process of the mapping function $H$. The $H$ can be learned from pseudo-noise label $\bar{n}_i$. Furthermore, the other constraint is to restrict the difference between pseudo-noise $\hat{x}_j-y_j$ and the output noise $H(\hat{x}_j)$}
   \label{fig:fig2}
\end{figure*}

\section{RELATED WORK}
\subsection{Conventional Neural Network Denoising}

Despite prior-based denoising had been widely used for many years, 
deep neural network has become popular in denoising tasks these days due to its superior performance. 
An earlier work by Jain \textit{et al.}~\cite{jain2008natural} introduced a CNN model for image denoising, and showed the representation powers through the visualization of hidden layers. 
Burger \textit{et al.}~\cite{burger2012image} proposed the multi-layer perceptron (MLP) model for denoising; however, it achieved similar or slightly better performance than BM3D over Gaussian noise. 
Typically, supervised learning of deep CNNs~\cite{zhang2017beyond,mao2016image,lefkimmiatis2017non,redcnn, sacnn, cagan} has shown superior performance over conventional image prior based approaches. 
In particular, DnCNN~\cite{zhang2017beyond} discovered that the combination of residual learning~\cite{he2016deep} and batch normalization~\cite{batch-norm} can greatly assist the training of CNNs for speeding up the training and boosting the denoising performance; however, it has a limitation with regard to the presence of noisy-residual (i.e., noise image) pairs. 
Recently, Lehtinen \textit{et al.}~\cite{noise2noise} introduced a Noise2Noise (N2N) method that can achieve similar performance employing only noisy pairs to the supervised learning performance. 
Even though N2N can overcome the requirement of clean-noisy pairs in the supervised learning, noise statistics is still an essential condition to generate noisy-noisy pairs. 

\subsection{Blind Denoising}
Blind denoising approaches~\cite{ulyanov2018deep,cbdnet,vdn,noise2self,noise2void,quan2020self2self,noise2sim,noise2kernel}
aim to restore noisy observations that are corrupted by unknown noise distribution without the supervision of clean targets. 
Deep Image Prior (DIP)~\cite{ulyanov2018deep} showed the usability of a hand-crafted prior, generated by a random-initialized neural network, for the image denoising task.
The internal image prior based approach is the early method of blind denoising. 
Recently, self-supervised learning based blind denoising approaches achieved the performance closed to that of supervised learning. 
N2S~\cite{noise2self} and N2V~\cite{noise2void} proposed a blind-spot scheme for training a CNN denoiser with only noisy images. 
Furthermore, they achieved significantly reduced deploying time through the external image prior. 
Blind denoising methods do not require clean-noisy pairs, but they still rely on the assumption of noise characteristics, such as zero-mean noise and pixel-wise signal-noise independence. 
More recently, S2S~\cite{quan2020self2self} successfully showed superior performance using internal image prior, that is, Bernoulli-sampled instances of only a single noisy image. Even though S2S is trained using a single noisy image, S2S outperforms external image prior based blind denoising methods.


\subsection{Unpaired Image Denoising}
To overcome the limitation of the blind denoising methods, unpaired image denoising methods~\cite{chen2018image,minh2019removing, park2019unpaired, wu2020unpaired,adn,drgan,adain} have gained much attention 
these days as a new denoising approach. 
Since the unpaired image denoising approaches can leverage the supervision of clean targets, zero-mean noise and pixel-wise signal independent assumptions are not prerequisite anymore. 
Furthermore, collecting of unpaired data is more feasible in a real setup, compared to using clean-noisy pairs. 
GCBD~\cite{chen2018image} demonstrated that the generative adversarial network (GAN)~\cite{goodfellow2014generative} can be trained to estimate the noise distribution from the noisy observations. However, it has a critical limitation: a zero-mean noise assumption. 
Quan \textit{et al.}~\cite{minh2019removing} proposed an asymmetrically cyclic adversarial network that consists of two generators. 
One generator can decompose a noisy observation to a clean-noise pair. 
The purpose of the other generator is to combine the clean-noise pair as a pseudo noisy image. 
The combination of two generators as an asymmetrical CycleGAN outperformed the state-of-the-art blind denoising methods without any image prior assumptions. However, it still has a limitation of requiring real noise image, which is often difficult to acquire. 
UIDNet~\cite{hong2020end} employed a conditional GAN (cGAN) to learn the noise distribution from noisy observations and generated clean-pseudo noisy pairs to train a denoiser. To secure the stability of training, they used the WGAN-GP~\cite{gulrajani2017improved} loss, that is, an improved version of WGAN~\cite{arjovsky2017wasserstein} with a gradient penalty. Furthermore, they proposed a sharpening technique that boosts the performance of the discriminator through the concatenation of input and filtered input. 
However, as shown in the following sections, using a simple generative  model  to  learn  the  noise  distribution  from examples is the main weakness of the method.
With the fusion of generative models and disentanglement networks, ADN~\cite{adn} and DRGAN~\cite{drgan} successfully reconstructed the noisy image with comparable performance to supervised learning methods in unpaired clean-noisy medical data. Especially, ADN adopted various encoders and decoders that conduct each allocated task to utilize the artifact disentanglement.

\begin{figure*}[t]
\centering
\includegraphics[width=16.5cm,keepaspectratio]{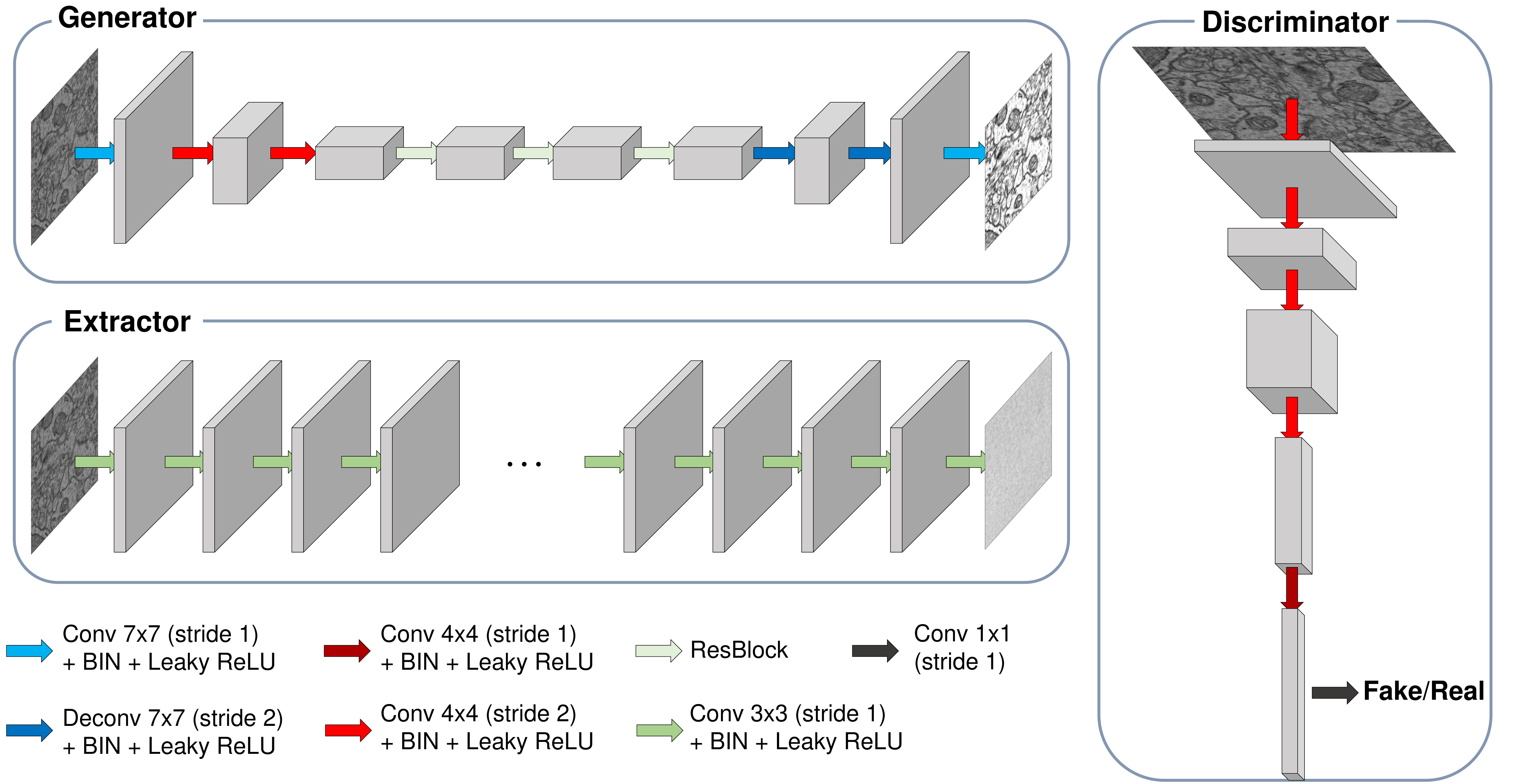}
\caption{Network structure: $F$ and $G$ employ generator structure. $H$ is an extractor. $D_X$ and $D_Y$ are discriminators.}
   \label{fig:fig3}
\end{figure*}
\section{METHOD}
\label{sec:method}
In this section, we introduce the details of ISCL. 
We focus on mapping between the noisy image domain $X$ and the clean image domain $Y$ using the two inter-domain mapping functions $F$ and $G$. 
Here, $F$ serves as a denoiser, and $G$ is the inverse of $F$, which is a noise generator (i.e., adding noise to the given clean image). 
To train $F$ and $G$, we employ $D_X$ and $D_Y$, which are discriminators, to distinguish a real sample and a fake sample (i.e., a domain shifted sample from another domain). 
However, adversarial losses are not sufficient constraints to train the discriminators for translating an ideal clean sample $y_i$ from a noisy sample $x_i$ due to the wide possible mapping space of $F(x_i)$. 
To generate a tighter mapping space from $x_i$, CycleGAN~\cite{zhu2017unpaired} proposed cycle consistency, i.e., $x\approx G(F(x))$ and $y\approx F(G(y))$, where $x\sim\mathcal{X}$ and $y\sim\mathcal{Y}$; $\mathcal{X}$ and $\mathcal{Y}$ are data distributions for the noisy observations and the clean sources, respectively. 
Therefore, we were faced with the problem that additional constraints are required to optimize $F$ and $G$ into bijective functions, i.e., a function for an ideal denoiser.

Suppose that $F$ and $G$ are bijective functions. Then, we can extract only a single noise image from $x_i$ by subtracting $F(x_i)$ from it. 
In other words, we can infer that there exists an injective function $H:X\rightarrow N$, where $N$ is a noise domain, that can extract the noise from the noisy observation. 
Based on this inference, we propose the cooperative learning concept to optimize the CycleGAN model and the noise extractor model simultaneously. 
Note that cooperative learning enables the training between the unpaired clean-noisy-based denoiser and the noise extractor based on self-supervision to boost the performance cooperatively.
In greater detail, five functions (i.e., $F$, $G$, $D_X$, $D_Y$, and $H$) are trained by assisting each other interdependently under the proposed novel losses generated by such assumptions in this paragraph. 
We denote the denoiser trained by Interdependent Self-Cooperative Learning ``ISCL".
\\
\subsection{Bypass-Consistency}
Here, we introduce the nested cycle consistency consisting of cycle-consistency and bypass-consistency. 
In Fig.~\ref{fig:fig2}b, we can find two mapping functions, $F$ and $G$, as generative models, trained by the following loss:
\begin{align}
    \mathcal{L}_{gen}(F,G,H,D_X,D_Y) &=  
    \mathcal{L}_{F}(F,D_Y) \nonumber \\
    &+\mathcal{L}_{G}(G,D_X) \nonumber \\
    &+\lambda\mathcal{L}_{nested}(F,G,H),
\end{align}
where $F$ translates a noisy target domain $X$ to a clean source domain $Y$ under the supervision of $D_Y$, and vice versa for $G$ and $D_X$. 
In detail, we borrow the generative loss based on hinge loss~\cite{lim2017geometric} to define $\mathcal{L}_{F}$ and $\mathcal{L}_{G}$ as follows:
\begin{align}
    \mathcal{L}_{F}(F,D_Y) &=  
    -\mathbb{E}_{x\sim\mathcal{X}}[D_Y(F(x))] \\
    \mathcal{L}_{G}(G,D_X) &=  
    -\mathbb{E}_{y\sim\mathcal{Y}}[D_X(G(y))]
\end{align}
and the nested cycle-consistency loss is defined as follows:
\begin{align}
    \mathcal{L}_{nested}(F,G,H) =  
    \mathcal{L}_{cycle}(F,G)+\mathcal{L}_{bypass}(F,H).
\end{align}
The cycle consistency loss $\mathcal{L}_{cycle}$ 
restricts the mapping space of $G(F(x))$ and $F(G(y))$, which is defined as follows:
\begin{align}
    \mathcal{L}_{cycle}(F,G) &= \mathbb{E}_{x\sim\mathcal{X}}||x-G(F(x))||_1 \nonumber \\
    &+\mathbb{E}_{y\sim\mathcal{Y}}||y-F(G(y))||_1.
\end{align}
Note that, even though the above cycle-consistency loss promotes bijections $F\circ G$ and $G\circ F$, there is no guarantee that both $F$ and $G$ are actually the bijective functions after convergence. 
Moreover, in unpaired domain translation tasks with cycle-consistency and generative adversarial models, performance degradation is occurred by self-adversarial attack~\cite{advattack} that the generator hides the output into an undetectable signal by the human eye or the discriminators. Even though the undetectable signal can be successfully reconstructed to the original signal, it affects the denoising performance. 
In other words, using only cycle-consistency is still insufficient for robust convergence of each function converged into the ideal function. 
%
If the injective function $H$ is available, then $\bar{y}$ is a pseudo-clean label for $x$, as shown in Fig.~\ref{fig:fig2} top. Then, we can restrict the mapping space of $F(x)$ into $\bar{y}$.
Moreover, we adopts the pseudo-noisy $\bar{x}$ to restrict the mapping space of $F(\bar{x})$ into $y$ real sample.
Finally, we propose the bypass-consistency to restrict the mapping space of the target denoiser $F$ through the pseudo label generated by $H$ as follows:
\begin{align}
    \mathcal{L}_{bypass}(F,H) &= \mathbb{E}_{x\sim\mathcal{X}}||F(x)-(x-H(x))||_1 \nonumber \\
    &+\mathbb{E}_{x\sim\mathcal{X},y\sim\mathcal{Y}}||y-F(y+H(x))||_1.
\end{align}
In other words, there exists two different approaches to mapping a noisy image $x$ into a clean source domain $Y$, either a bijective function $F$ or an injective function $H$, as shown in Fig.~\ref{fig:fig2}b. 
The bypass-consistency promotes two outputs generated by the two mapping functions $F$ and $H$ to be similar to each other to satisfy our assumption. 
In addition, as shown in Fig.~\ref{fig:fig2} bottom, the reconstructed outputs $\tilde{y}_j$ and $\hat{y}_j$ generated from real clean source $y_j$ through $F(G(y_j))$ and $F(y_j+H(x_i))$ should be similar to the clean source $y_j$. 
In summary, we introduced the nested cycle-consistency to cooperate between the generators of CycleGAN and the noise extractor $H$ under the supervision of discriminators $D_X$ and $D_Y$.



\begin{algorithm}[t]
\SetAlgoLined
\algorithmicrequire{$\lambda=30$ for $\mathcal{L}_{nested}$, $\gamma=0.5$, $N_{epoch}$, $N_{iter}$, batch size m, patch size of $64\times64$}\\
\algorithmicrequire{$N_{swa}$, cycle length $c$, synchronization period $k$, weights step size $\alpha$}\\
\algorithmicrequire{Initialize parameters $\theta^{(F)}$, $\theta^{(G)}$, $\theta^{(H)}$ $\theta^{(D_X)}$, $\theta^{(D_Y)}$}\\
$\phi^{(D_X)} \gets \theta^{(D_X)}, \phi^{(D_Y)} \gets \theta^{(D_Y)}, \phi^{(H)} \gets \theta^{(H)}$\\
\For{$e=0,...,N_{epoch}$}
{
\If{$e\geq N_{swa}$}{
$\phi^{(F)}\gets \theta^{(F)}, \phi^{(G)}\gets \theta^{(G)}$
}
\For{$t=1,...,N_{iter}$}
{ Unpaired mini-batch of noisy image patches $\{x^{(i)}\}^m_{i=1}$, and clean image patches $\{y^{(j)}\}^m_{j=1}$ from data generating distribution $\mathcal{X}$ and $\mathcal{Y}$ feed into each loss. \\
Update $F, G$: $\theta^{(F)},\theta^{(G)}\gets$ Radam \\
$(\nabla_{\theta^{(F)},\theta^{(G)}} \mathcal{L}_{gen}({F,G,H,D_X,D_Y}), \theta^{(F)},\theta^{(G)})$ \\
\If{$e\geq N_{swa}$}{
   \If{$mod(t+(e-N_{swa})*N_{iter},c) = 0$}{
   $n_{models} \gets \frac{t+(e-N_{swa})*N_{iter}}{c}$  \\
   $\phi^{(F)}\gets \frac{\phi^{(F)}\cdot n_{models}+\theta^{(F)}}{n_{models}+1}$ \\
   $\phi^{(G)}\gets \frac{\phi^{(G)}\cdot n_{models}+\theta^{(G)}}{n_{models}+1}$
   }
   }
Update $D_X, D_Y$: $\theta^{(D_X)},\theta^{(D_Y)}\gets$ Radam(\\
$\nabla_{\theta^{(D_X)},\theta^{(D_Y)}}\mathcal{L}_{dis}({F,G,H,D_X,D_Y}), \theta^{(D_X)},\theta^{(D_Y)})$ \\
Update $H$: $\theta^{(H)} \gets$ Radam( \\
$\nabla_{\theta^{(H)}}\mathcal{L}_{self}({F,G,H}), \theta^{(H)})$ \\
\If{$mod(t+e*N_{iter},k) = 0$}{
$\phi^{(D_X)} \gets \phi^{(D_X)}+ \alpha(\theta^{(D_X)}-\phi^{(D_X)})$ \\
$\phi^{(D_Y)} \gets \phi^{(D_Y)}+ \alpha(\theta^{(D_Y)}-\phi^{(D_Y)})$ \\
$\phi^{(H)} \gets \phi^{(H)}+
\alpha(\theta^{(H)}-\phi^{(H)})$ \\ 
$\theta^{(D_X)} \gets \phi^{(D_X)}$,
$\theta^{(D_Y)} \gets \phi^{(D_Y)}$,
$\theta^{(H)} \gets \phi^{(H)}$
}
}
}
\algorithmicreturn{ $\phi^{(F)}, \phi^{(G)}, \phi^{(D_X)}, \phi^{(D_Y)}, \phi^{(H)}$}
\caption{Interdependent Self-Cooperative Learning Algorithm}
\label{algo:ISCL}
\end{algorithm}

\subsection{Boosting Discriminators}
Discriminators use real and fake samples to optimize the model based on the adversarial losses. 
In conventional adversarial learning, discriminators $D_X$ and $D_Y$ depend on only fake samples generated by $F$ and $G$. 
To improve the ability of discriminators, the fake samples generated by $H$ also have the advantage of the cooperative learning. 
We propose an additional boosting loss to improve the discriminator's capability to distinguish fake samples as follows:
\begin{align}
    \mathcal{L}_{dis}(F,G,H,D_X,D_Y) &= 
    \mathcal{L}_{D_{Y}}(F,D_Y) \nonumber \\
    &+\mathcal{L}_{D_{X}}(G,D_X) \nonumber \\
    &+\mathcal{L}_{bst}(H,D_X,D_Y).
\end{align}
For the discriminators, we employ hinge loss~\cite{lim2017geometric} to train the adversarial network against the generators, $F$ and $G$ as follows:
\begin{align}
    \mathcal{L}_{D_{Y}}(F,D_Y) &= \mathbb{E}_{y\sim\mathcal{Y}}[min(0,1-D_Y(y))] \nonumber \\
    &+\mathbb{E}_{x\sim\mathcal{X}}[min(0,D_Y(F(x))] \nonumber \\
    \mathcal{L}_{D_{X}}(G,D_X) &= \mathbb{E}_{x\sim\mathcal{X}}[min(0,1-D_X(x))] \nonumber \\
    &+\mathbb{E}_{y\sim\mathcal{Y}}[min(0,D_X(G(y))] 
\end{align}
and the boosting loss is defined with additional fake samples generated by $H$ as follows:
\begin{align}
    \mathcal{L}_{bst}(H,D_X,D_Y) &=  \mathbb{E}_{x\sim\mathcal{X}}[min(0,D_Y(x-H(x)))] \nonumber \\
    &+\mathbb{E}_{x\sim\mathcal{X},y\sim\mathcal{Y}}[min(0,D_X(y+H(x))].
\end{align}
$\mathcal{L}_{bst}$ promotes the ability to discriminate fake clean $\bar{y}$ and fake noisy $\bar{x}$ using a noise $H(x)$, as shown in Fig.~\ref{fig:fig2}c.
The discriminators are interdependently optimized by the outputs of generators and the noise extractor with real unpaired data.

\subsection{Pseudo-Noise Label}
The basic concept of self-residual learning is to construct a pseudo-noise label from CycleGAN for training the noise extractor. 
In the next step, the noise extractor $H$ will assist the training of CycleGAN to boost the performance. 
We express the mapping function $H$ as the noise extractor, as shown in  Fig.~\ref{fig:fig2}d. 
If $F$ is a bijective function, then we can generate a unique noise map $n$ by $x-F(x)$. 
In other words, we employ the pseudo-noise label $\bar{n}$ generated by $x-F(x)$ to learn the capability of the noise extraction.
Using this pseudo-noise label, we can optimize the mapping function $H$ by the following loss:
\begin{align}
    \mathcal{L}_{pseudo}(F,H) =  
    \mathbb{E}_{x\sim\mathcal{X}}||H(x)-(x-F(x))||_1.
\end{align}
In addition, we can generate the single noise $n$ by $G(y)-y$ if $G$ is also a bijective function. Moreover, $H(G(y))$ can extract the same noise map $n$ because of the injective function assumption for $H$. 
To reduce (constrain) the mapping space of the $H(\hat{x})$, we add the noise-consistency loss as follows:
\begin{align}
    \mathcal{L}_{nc}(G,H) =  
    \mathbb{E}_{y\sim\mathcal{Y}}||G(y)-y-H(G(y))||_1.
\end{align}
Finally, we can optimize $H$ function with the following loss:
\begin{align}
    \mathcal{L}_{self}(F,G,H) = \mathcal{L}_{pseudo}(F,H) + \mathcal{L}_{nc}(G,H).
\end{align}
$\mathcal{L}_{self}$ is a self-supervision based loss because it utilizes each sample $x$ or $y$ even if $x$ and $y$ are unpaired. In other words, the self-residual learning through $\mathcal{L}_{self}$ can be applicable to the task in which unpaired data are available. 
The self-residual learning with $\mathcal{L}_{self}$ leads to stable convergence and performance improvement similar to co-teaching scheme~\cite{han2018co}.
Algorithm~\ref{algo:ISCL} is the pseudo-code of ISCL where stochastic weight averaging (SWA)~\cite{SWA} and Lookahead~\cite{lookahead} schemes are used with the RAdam~\cite{liu2019variance} optimizer for optimal training.
%
The final denoising output of ISCL is an ensemble of outputs $F$ and $H$ as follows:
\begin{align}
    y = \gamma F(x) + (1-\gamma)(x-H(x))
\end{align}
where $0 \leq \gamma \leq 1$. We used $\gamma = 0.5$ in our experiments.

\begin{figure*}[t]
\includegraphics[width=18cm,keepaspectratio]{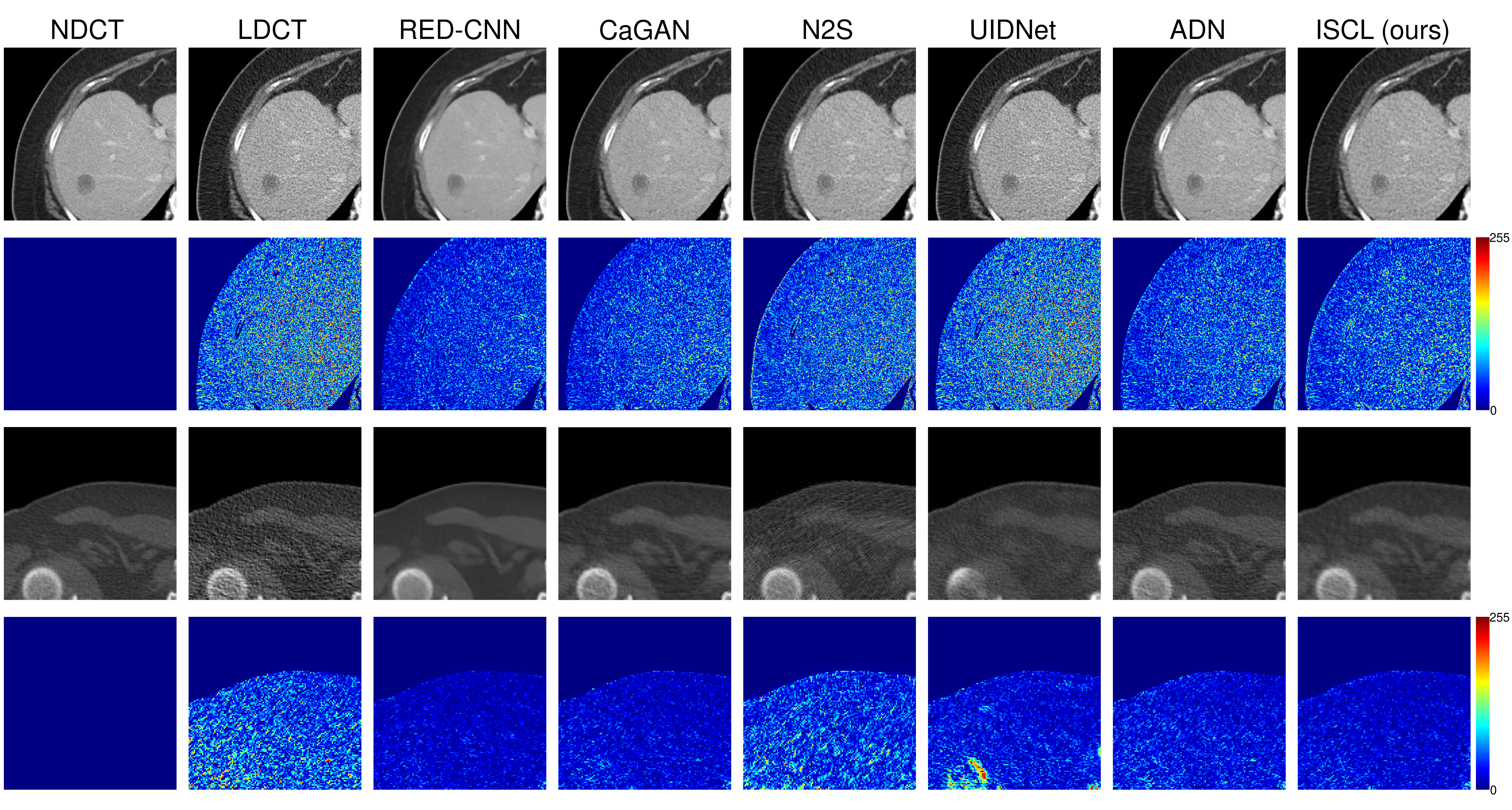}
\caption{Qualitative performance of supervised denoisers (i.e. RED-CNN and CaGAN), a blind denoiser (i.e. N2S), and unpaired image denoisers (i.e. UIDNet, ADN, and ISCL) on low-dose CT data. First row indicates the denoising results of the small portion of abdominal CT that are normalized under [-160, 240] Hounsfield Unit (HU). Third row shows the denoising results of the small portions of chest CT that are normalized under [-400, 1000] HU. Second and fourth rows are the error heat mpas showing the difference between NDCT and each result.}
   \label{fig:fig4}
\end{figure*}
\section{EXPERIMENTS}
\label{sec:experiments}
In this section, we demonstrate the performance of ISCL via quantitative and qualitative evaluation on synthetic and real EM datasets and a low-dose CT dataset. 
For the CT dataset, we also conducted ablation study to elaborate how each loss contributes to the performance of the method.
Our experiments consists of three parts: (1) Ablation study and performance assessment on the 2016 NIH-AAPM-Mayo Clinc Low Dose CT Grand Challenge dataset~\cite{LDCT}; (2) Quantitative performance evaluation on synthetic noisy EM image generated by adding film noise and charge noise into clean EM images~\cite{minh2019removing}; and (3) Qualitative performance comparison on real EM images corrupted with film noise and charge noise in which the ground-truth clean images are not available~\cite{minh2019removing}. 
For the fair comparison with other methods, we 
used the source code provided by the authors (downloaded from their website) with the optimal hyper-parameters 
empirically found for the best performance or the best parameters reported by the authors. %
\subsection{Implementation Details}
We construct five deep neural networks, generators $F$ and $G$, discriminators $D_X$ and $D_Y$, and noise extractor $H$, to train the ISCL denoiser. 
All architectures are illustrated in Fig.~\ref{fig:fig3}. The noise extractor $H$ is adopted from DnCNN~\cite{zhang2017beyond} except the normalization method. 
We replace the batch normalization~\cite{batch-norm} layers with Batch-Instance normalization~\cite{nam2018batch} layers that can have advantages of batch normalization and instance normalization~\cite{ulyanov2016instance}; it preserves useful textures while selectively normalizing only disturbing textures.
%
As shown in Fig.~\ref{fig:fig3}, we adopt a fully convolutional network architecture~\cite{long2015fully} to handle different input sizes. 
We randomly extract patches of size $64\times64$ to increase the batch size to fit to the limited GPU memory size.
%
Each mini-batch contains randomly selected patches from unpaired clean and noisy images.
%
As shown in Algorithm~\ref{algo:ISCL}, the three RAdam~\cite{liu2019variance} optimizers are used to train the generators, the discriminators, and the extractor.
\begin{figure*}[t]
\includegraphics[width=18cm,keepaspectratio]{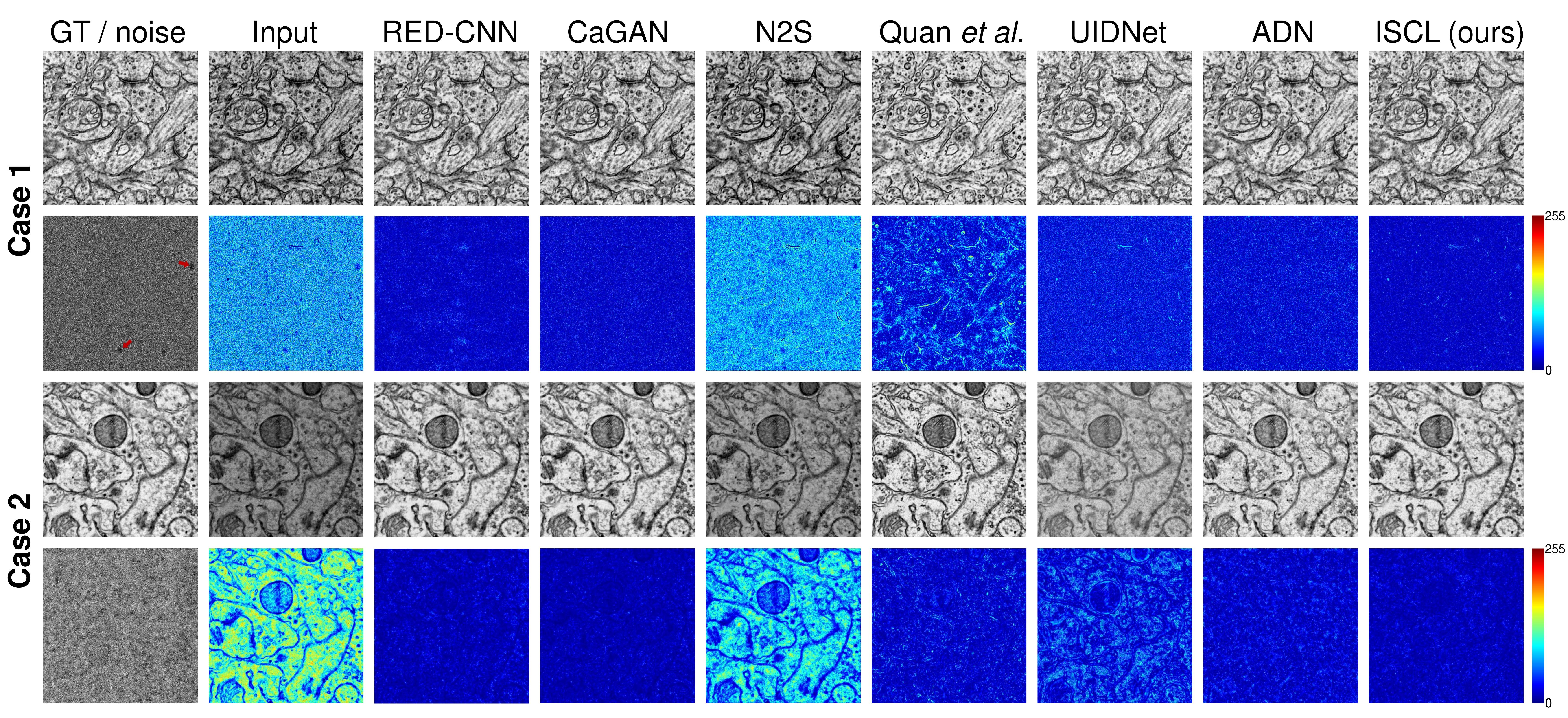}
\caption{Comparison with state-of-the-art denoising methods on synthetic noisy EM data in case 1 (charge noise) and 2 (film noise). Second and fourth rows are the error heat map showing the difference between the ground-truth and the result.}
   \label{fig:fig5}
\end{figure*}
Furthermore, since self-cooperative learning is sensitive to each other performance at each iteration, we empirically found the suitable generalization methods for each network architecture. 
We employ a SWA~\cite{SWA} for generalization of generators to avoid unstable convergence since the unstable performance at each iteration negatively affects the self-cooperative learning. 
We also employ the Lookahead~\cite{lookahead} generalization scheme to improve the learning stability for discriminators $D_X$ and $D_Y$, and noise extractor $H$. 
The learning rate is initially set to 1e-4, and is linearly decayed up to 1e-6 for all RAdam optimizers.
\subsection{Low-Dose CT Denoising}
For performance evaluation on low-dose CT, we used the abdominal and chest CT images in the 2016 NIH-AAPM-Mayo Clinc Low Dose CT Grand Challenge dataset~\cite{LDCT}. 
%
%
In this dataset, LDCT and normal dose CT (NDCT) indicate the noisy and clean images, respectively. 
We randomly selected 30 anonymous patients for training and 10 anonymous patients for testing in the abdominal and chest datasets. 
We collected 2944 and 1433 slices (each is of size 512$\times$512 pixels) for training and testing in the abdominal case, respectively. 
For the chest case, we randomly selected 3000 slices from among 6687 training images to reduce the training cost, and collected 3254 slices from 10 anonymous patients for testing. %
We compared ISCL with supervised denoisers (i.e., DnCNN~\cite{zhang2017beyond}, RED-CNN~\cite{redcnn}, and CaGAN~\cite{cagan}), blind denoisers (i.e., BM3D~\cite{dabov2007}, N2S~\cite{noise2self}, and N2V~\cite{noise2void}), and unpaired image denoisers (i.e., UIDNet~\cite{hong2020end} and ADN~\cite{adn}).
\begin{table}[t]
\centering
\renewcommand{\arraystretch}{1.3}
\renewcommand{\tabcolsep}{6.5pt}
\caption{Performance comparison on low-dose CT dataset. P.S.: Paired Supervision, B.S.: Blind Supervision, U.S.: Unpaired Supervision, (a) CycleGAN only, (b) Self-Residual Network trained by pseudo-noise label, (c) $\mathcal{L}_{bypass}$, (d) $\mathcal{L}_{bst}$, (e) $\mathcal{L}_{nc}$. Bottom-most results are final performance of the proposed method ISCL. The best PSNR in each case except P.S. is highlighted in bold.}
\label{tab:tab1}
\begin{tabular}{||c||c|c|c|c|c||}
\hline
\multirow{2}{*}{Type} & \multirow{2}{*}{Method} & \multicolumn{2}{c|}{Abdominal} & \multicolumn{2}{c||}{Chest}  \\
     &        & PSNR & SSIM & PSNR & SSIM  \\ \hline 
\multirow{3}{*}{P.S.} & DnCNN~\cite{zhang2017beyond} & 30.57 & 0.8192 & 27.47 & 0.7354 \\ \cline{2-6}
                      & RED-CNN~\cite{redcnn} & 33.09 & 0.9075 & 29.54 & 0.7904 \\ \cline{2-6}
                      & CaGAN~\cite{cagan} & 31.98 & 0.8968 & 27.69 & 0.7766 \\ \hline \hline 
\multirow{3}{*}{B.S.}& BM3D~\cite{dabov2007}
& 30.31 & 0.8730 & 26.75 & 0.7336\\ \cline{2-6}
                     & N2S~\cite{noise2self} 
                     & 28.94 & 0.8355 & 23.76 & 0.6672 \\ \cline{2-6}
                     & N2V~\cite{noise2void} 
                     & 28.32 & 0.7961 & 26.30 & 0.7283  \\ \hline
\multirow{8}{*}{U.S.}& UIDNet~\cite{hong2020end} 
& 28.91 & 0.8470 & 24.15 & 0.7221  \\ \cline{2-6}
                     & ADN~\cite{adn} 
& 30.41 & \textbf{0.8857} & 26.79 & 0.7573  \\ \cline{2-6}
                     & (A) & 22.33 & 0.7561 & 22.06 & 0.6236 \\ \cline{2-6} 
                     & (A)+(B) & 22.10 & 0.7954 & 22.58 & 0.5815  \\ \cline{2-6}
                     & (A)+(B)+(C) & 29.43 & 0.8811 & 26.61 & 0.7533 \\ \cline{2-6}
                     & (A)+(B)+(C)+(D) & 30.13 & 0.8819 & 26.89 & 0.7569  \\ \cline{2-6}
                     & (A)+(B)+(C)+(D)+(E) & \textbf{30.61} & 0.8849 & \textbf{26.93} & \textbf{0.7587} \\ \hline
\end{tabular}
\end{table}
\begin{table}[t]
\centering
\renewcommand{\arraystretch}{1.3}
\renewcommand{\tabcolsep}{3.5pt}
\caption{Specifications for our EM experiment cases.}
\begin{tabular}{||c|c|c|c||}
\hline
Case & Noise-Free Images & Noise Types & Noisy Images (Scenario) \\ \hline
1 & $TEM_{ZB}$ & Charge & $TEM_{ZB}$ + Charge (Synthetic) \\ \hline
2 & $TEM_{DR5}$ & Film & $TEM_{DR5}$ + Film (Synthetic) \\ \hline
3 & $TEM_{ZB}$ & Charge & $SEM_{ZB}$ (Real) \\ \hline
4 & $TEM_{DR5}$ & Film & $TEM_{PPC}$ (Real) \\ \hline
\end{tabular}
\label{tab:tab2}%
\end{table}%
\begin{table}[t]
\centering
\renewcommand{\arraystretch}{1.3}
\renewcommand{\tabcolsep}{8pt}
\caption{P.S.: Paired Supervision, B.S.: Blind Supervision, U.S.: Unpaired Supervision. Quantitative PSNR and SSIM results on case 1 and 2. The best PSNR in each case except P.S. is highlighted in bold.}
\label{tab:tab3}
\begin{tabular}{||c||c|c|c|c|c||}
\hline
\multirow{2}{*}{Type} & \multirow{2}{*}{Method} & \multicolumn{2}{c|}{Charge noise} & \multicolumn{2}{c||}{Film noise}  \\
     &        & PSNR & SSIM & PSNR & SSIM \\ \hline 
\multirow{3}{*}{P.S.}     & DnCNN~\cite{zhang2017beyond}
& 28.27 & 0.9172 & 27.55 & 0.8964  \\ \cline{2-6}
                          & RED-CNN~\cite{redcnn}
& 28.61 & 0.9230 & 28.02 & 0.9049  \\ \cline{2-6}
                          & CaGAN~\cite{cagan}
& 28.60 & 0.9186 & 28.03 & 0.9020 \\ \hline \hline 
\multirow{3}{*}{B.S.}     & BM3D~\cite{dabov2007} 
& 17.85 & 0.7873 & 12.85 & 0.6097 \\ \cline{2-6}
                          & N2S~\cite{noise2self} 
                          & 18.75 & 0.8680 & 13.47 & 0.7942  \\ \cline{2-6}
                          & N2V~\cite{noise2void}
                          & 18.06 & 0.8286 & 12.86 & 0.6860  \\ \hline
\multirow{3}{*}{U.S.}     & Quan \textit{et al.}~\cite{minh2019removing}
                          & 22.32 & 0.8785 & 23.44 & 0.8288  \\ \cline{2-6} 
                          & UIDNet~\cite{hong2020end} 
                          & 23.11 & 0.8592 & 21.34 & 0.7826 \\ \cline{2-6} \cline{2-6}
                          & ADN~\cite{adn}
                          & 25.67 & 0.8686 & 24.37 & 0.8535 \\ \cline{2-6}
                          & ISCL (ours) & \textbf{27.12} & \textbf{0.9054} & \textbf{27.06} & \textbf{0.8915} \\ \hline
\end{tabular}
\end{table}
\noindent For the blind denoising methods, all LDCT slices of the training set are used to train N2S and N2V models without NDCT.
\begin{figure*}[t]
\centering
\includegraphics[width=18cm,keepaspectratio]{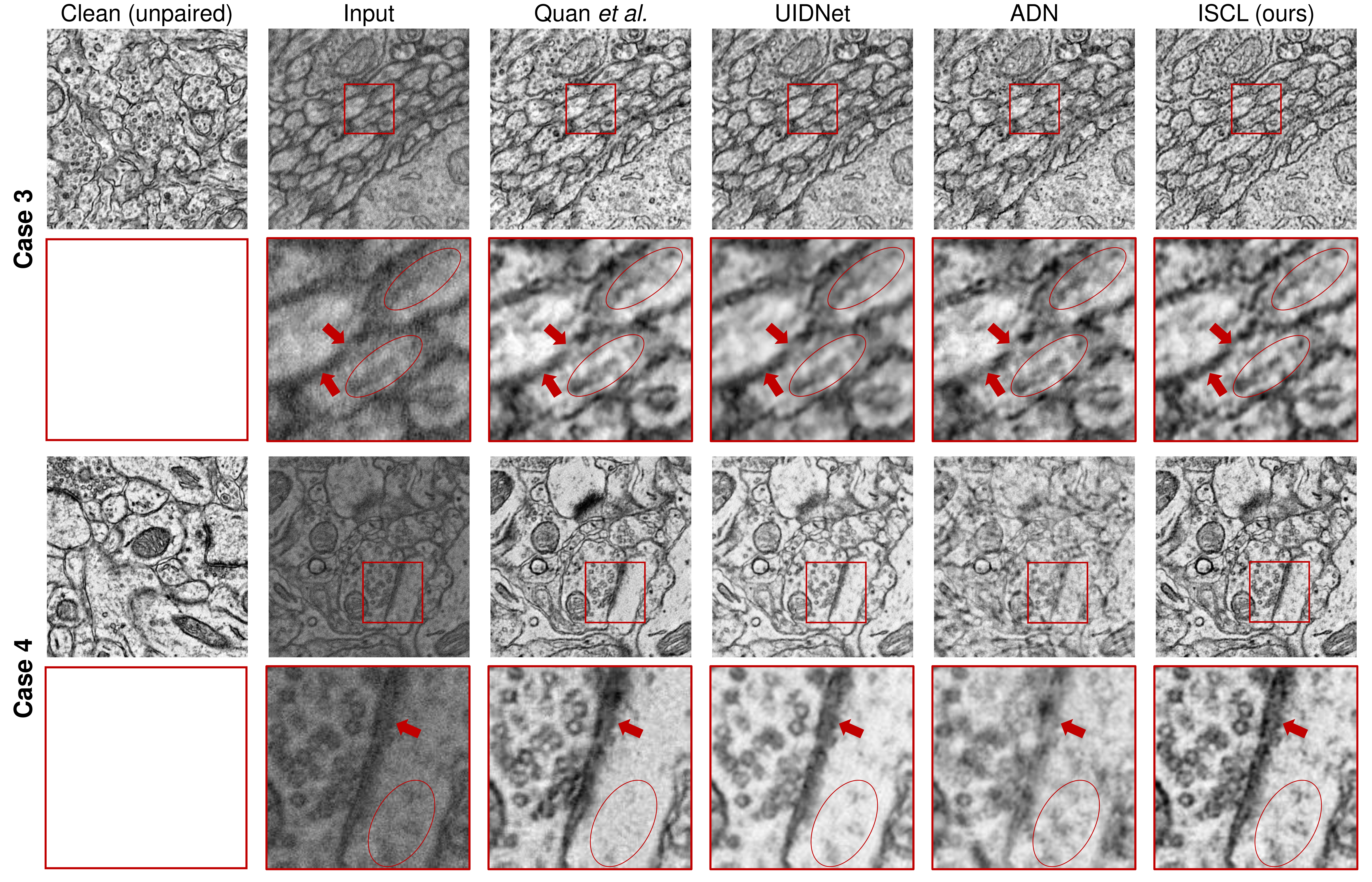}
\caption{Qualitative assessment of the denoising quality without using ground truth on real EM images corrupted with charge noise (case 3) and film noise (case 4). Note that the proposed method enhances the quality of the input noisy images comparable to the clean TEM images (shown on the left) without paired clean ground truth images.}
   \label{fig:fig6}
\end{figure*}
Unlike the supervised learning and the blind denoising methods, unpaired image denoising approaches (i.e., UIDNet, ADN, and ISCL) require unpaired clean-noisy data.
Therefore, we divided the data into two non-overlapping groups; one group contains only NDCT as a clean dataset, the other group contains only LDCT as a noisy dataset. \noindent
Table~\ref{tab:tab1} and Fig.~\ref{fig:fig4} show the quantitative and qualitative comparison of the results, respectively. 
The supervised denoiser RED-CNN tends to over-smooth the image (note the cartoon-like, a piecewise linear resulting image of RED-CNN in Fig.~\ref{fig:fig4}) but the quantitative results are better than the other non-supervised methods (B.S. and U.S. in Table~\ref{tab:tab1}). 
ISCL and CaGAN generate qualitatively better results than RED-CNN (see the first and third rows of Fig.~\ref{fig:fig4}, image textures of ISCL and CaGAN are closer to NDCT) while quantitatively comparable to RED-CNN (similar or higher PSNR and SSIM).
When comparing with blind (BM3D, N2S, and N2V) and unpaired image denoisers (UIDNet and ADN), ISCL outperformed them in PSNR and SSIM (ADN was slightly better in SSIM for the abdominal data, only about 0.0008, which seems statistically insignificant). 
The error heat map in Fig.~\ref{fig:fig4} also shows that ISCL's results are much less noisier than the others. 
The bottom five rows in Table~\ref{tab:tab1} are the result of ablation study to assess the effect of the proposed loss functions. 
The baseline is a vanilla CycleGAN (A), and we incrementally added each loss term (B, C, D and E) and measured PSNR and SSIM. 
We observed that including more losses ($\mathcal{L}_{bypass}$, $\mathcal{L}_{bst}$, $\mathcal{L}_{nc}$) always lead to better performance. 
Especially, we observed a sharp performance jump when $\mathcal{L}_{bypass}$ is added, showing the dominant performance gain was from the bypass consistency; this is the core idea of cooperative learning where $F$ and $H$ are jointly promoting the performance of each method. 
\subsection{Synthetic Noisy EM Denoising}
For quantitative assessment, we used synthetically generated noisy EM images.
We used the same dataset of charge noise and film noise first used in Quan \textit{et al.}~\cite{minh2019removing}, as listed in Table~\ref{tab:tab2}. 
We used 128 images of $512\times 512$ for each type of noise free ($TEM_{ZB}$ and $TEM_{DR5}$) and noisy (synthetically generated) images, listed as case 1 and 2. 
$TEM_{ZB}$ and $TEM_{DR5}$ are noise-free clean TEM images of a juvenile zebrafish brain and a mouse brain respectively, and the corresponding noisy images are synthetically generated by adding a charge noise (for $TEM_{ZB}$) or multiplying a film noise (for $TEM_{DR5}$). 
The example noise images are shown in Fig.~\ref{fig:fig5} (under the ground truth image).
To avoid test set selection bias, we ran 4-fold cross validation (3 to 1 split) where each test set consists of 32 images.
To compensate the small size of EM training set, we applied rotation and mirroring data augmentation. 
%
%
As shown in the first row of Fig.~\ref{fig:fig5}, we observed that N2S fails to recover the correct brightness due to the non-zero mean noise distribution. 
We also observed in the error heat map of case 1 that N2S 
did not remove structural noise well. 
As shown in Tab.~\ref{tab:tab3}, the other blind denoising methods (BM3D and N2V) also performed poorly. 
As for unpaired denoising cases, the result of Quan \textit{et al.} on case 1 shows strong errors near the edges. 
UIDNet and ADN also show Gaussian noise-like corruption in the result of case 1. For case 2, we discovered that the shape-dependent noise in the result of UIDNet.
Unlike the other unpaired image denoising methods, ISCL successfully restores the structure noise with correct brightness in case 1 and case 2 of Fig.~\ref{fig:fig5}.
In addition to qualitative results, ISCL outperforms all comparison methods except the supervised learning in Table~\ref{tab:tab3}. Furthermore, ISCL achieves PSNRs $>$ 27dB that is the highest values among  all the unpaired denoising methods compared with. 
Consequently, it is clearly shown that ISCL can effectively eliminate unconventional noise corruption via training using only unpaired data without noise distribution prior.
\begin{figure}[t]
\centering
\includegraphics[width=8.5cm,keepaspectratio]{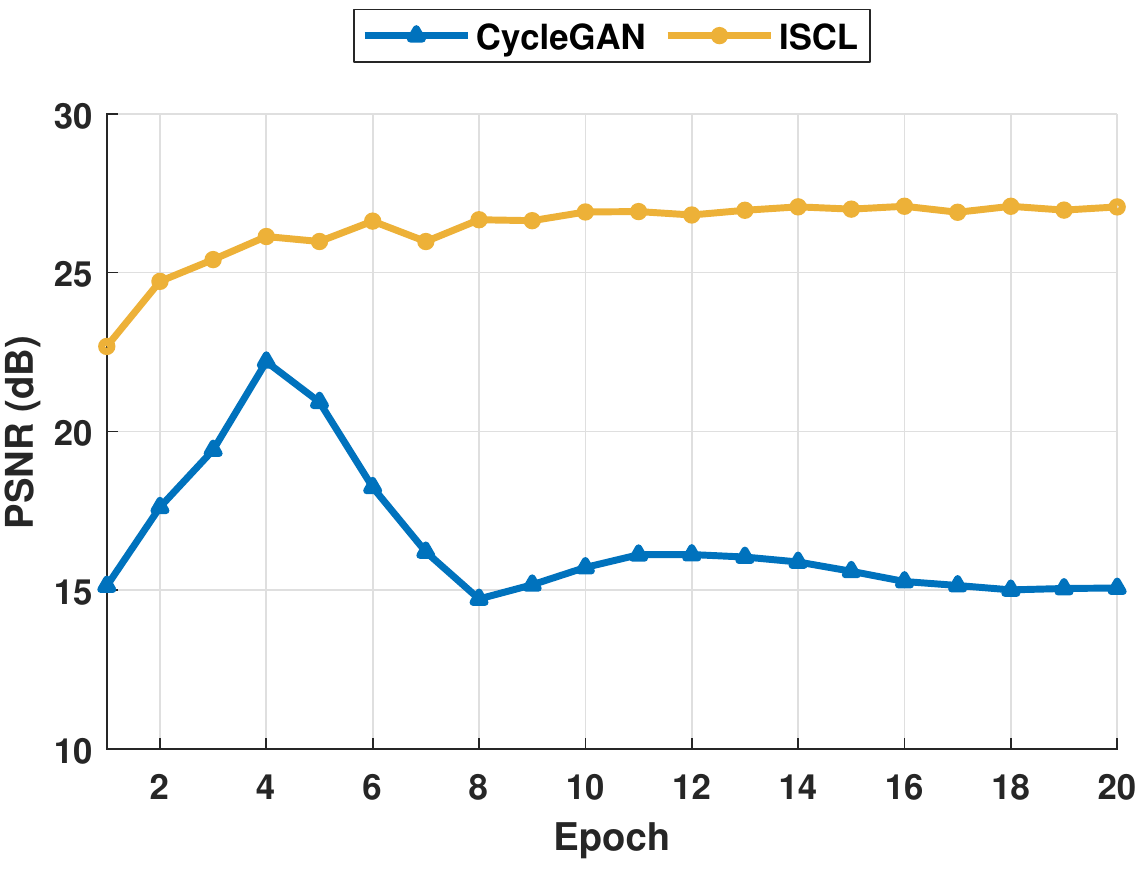}
\caption{An example graph for validation data of film noise (case 2); the validation data is also unseen data during training. In this graph, the performance of ISCL indicates the validation PSNR of $F(x)$ instead the ensemble $(F(x)+x-H(x))*0.5$ for a fair comparison.}
   \label{fig:fig7}
\end{figure}

\subsection{Real Noisy EM Denoising}

To assess the performance of the proposed method in a realistic setup, we compared the denoising quality on $SEM_{ZB}$ (case 3) and $TEM_{PPC}$ (case 4), which are real noisy EM images corrupted with charge noise and film noise, respectively. 
In this experiment, we used clean EM images ($TEM_{ZB}$ and $TEM_{DR5}$) as unpaired clean images to train unpaired denoising methods. 
We already observed in the previous synthetic noise removal experiment that blind denoising methods performed poorly on film and charge noise. Therefore, we tested only unpaired denoising methods in this experiment. 

In the absence of ground truth data, we can only assess the visual quality. 
Fig.~\ref{fig:fig6} shows the qualitative results for case 3 and 4. 
All four unpaired image denoising methods we compared (Quan~\textit{et al.}, UIDNet, ADN, and ISCL) successfully restored the image contrast similar to the unpaired clean images used for training.
%
%
Although the difference seems subtle, Quan~\textit{et al.}, UIDNet, and ADN sometimes over-smooth or over-enhance the image compared to ISCL. 
For example, in case 3, ISCL restored cell membranes and particles much clearer compared to the other methods (see the red arrows and circles). 
In case 4, Quan~\textit{et al.} restored synapse and vesicles well but small particles are removed. 
UIDNet and ADN over-smoothed the image and the results look fuzzy and blurry. 
ISCL showed the best result among all in case 4 where all cellular neural structures are well restored. 
In summary, ISCL demonstrated similar or better denoising quality compared to Quan~\textit{et al.}, while outperforming UIDNet and ADN in terms of overall image contrast and feature details for real EM image denoising.

\subsection{Discussion}

In the previous section, we demonstrated how the proposed constraints contribute to the performance of ISCL (Table~\ref{tab:tab1}). 
%
To further analyze the main source of the performance of ISCL, we compare the validation performance graph of a vanilla CycleGAN and ISCL (Fig.~\ref{fig:fig7}).
%
In this experiment, we used the same generator architecture for both methods; the only difference is that ISCL is trained using the proposed self-cooperative learning scheme. 
As shown in Fig.~\ref{fig:fig7}, the vanilla CycleGAN without the self-cooperative learning showed unstable performance; furthermore, it converged to lower validation performance even though SWA and Lookahead were applied to generators and discriminators. 
However, ISCL showed stable performance with higher PSNR for the validation data. Moreover, it reached the level of the maximum PSNR of CycleGAN even if each generator $F$ of CycleGAN and ISCL for denoising has the same structure, as shown in Fig.~\ref{fig:fig3}. 
We conclude that self-cooperative learning closely leads to a global optimal point under the same conditions, such as the number of parameters and training data. 
\begin{figure}[t]
    \centering
    \includegraphics[width=0.5\textwidth]{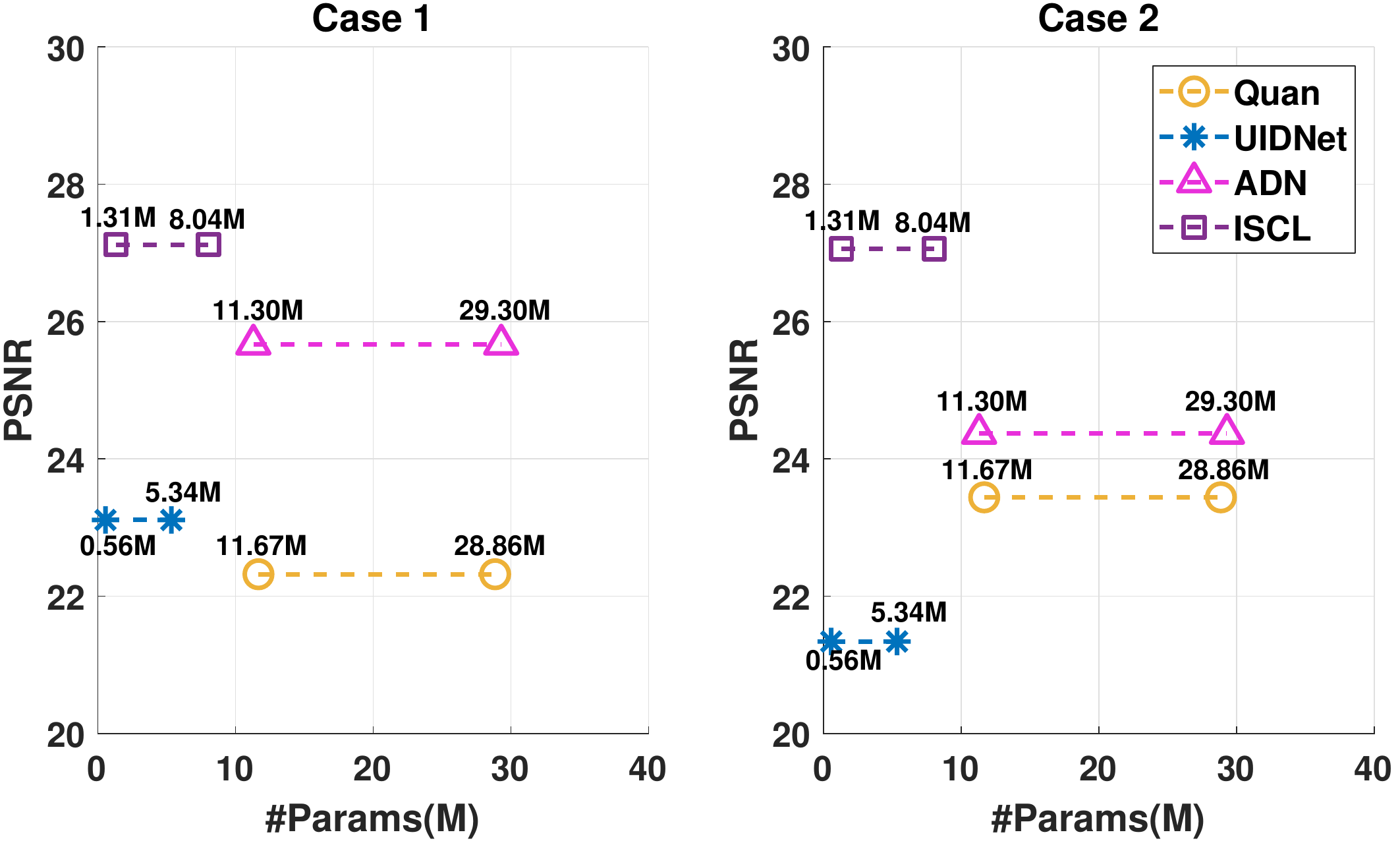}
    \caption{Performance comparison in terms of model size for case 1 (left) and 2 (right) in EM image denoising. Each model is represented by two points connected with a dotted line; the left point indicates the number of trained variables for deployment, and the right point indicates the number of trainable variables for training.}
    \label{fig:fig8}
\end{figure}

\begin{table*}[t]
\centering
\renewcommand{\arraystretch}{1.3}
\renewcommand{\tabcolsep}{8pt}
\caption{P.S.: Paired Supervision, U.S.: Unpaired Supervision. Quantitative PSNR and SSIM results on different dose-levels LDCT. The best PSNR and SSIM in each case except P.S. is highlighted in bold.}
\label{tab:tab5}
\begin{tabular}{||c||c|c|c|c|c|c|c|c|c||}
\hline
\multirow{2}{*}{Type} & \multirow{2}{*}{Method} & \multicolumn{2}{c|}{$I_0=1*10^4$} & \multicolumn{2}{c|}{$I_0=5*10^4$} &
\multicolumn{2}{c|}{$I_0=1*10^5$} &
\multicolumn{2}{c||}{$I_0=5*10^5$}  \\
     &        & PSNR & SSIM & PSNR & SSIM & PSNR & SSIM & PSNR & SSIM \\ \hline 
\multirow{1}{*}{P.S.}     & RED-CNN~\cite{redcnn}
& 19.36 & 0.6801 & 24.38 & 0.8053 & 27.18 & 0.8441 & 30.15 & 0.8914 \\ \hline \hline 
\multirow{3}{*}{U.S.}     & UIDNet~\cite{hong2020end} 
                          & 15.61 & 0.6103 & 21.21 & 0.6947 & 23.89 & 0.7463 & 28.76 & 0.8534\\ \cline{2-6} \cline{2-10}
                          & ADN~\cite{adn}
                          & 15.62 & \textbf{0.6578} & 21.47 & \textbf{0.7676} & \textbf{25.63} & \textbf{0.8204} & \textbf{29.21} & \textbf{0.8828}\\ \cline{2-10}
                          & ISCL (ours) & \textbf{18.76} & 0.6326 & \textbf{23.61} & 0.7273 & \textbf{25.63} & 0.7745 & 28.79 & 0.8664 \\ \hline
\end{tabular}
\end{table*}

For a fair comparison of model performance, we compared the accuracy versus the model size (Fig.~\ref{fig:fig8}).
In this plot, the horizontal axis represents the model size and the vertical axis represents the accuracy in PSNR (higher the better). 
Each model is represented using two points connected by a dotted line; the left point is for the deployment, and the right point is for the training (note that the model used during training is usually larger due to multiple generators and adversarial modules).
Therefore, the models in the upper left region are optimal ones in terms of the model size and accuracy. 
ISCL used about nine times smaller trained variables than Quan~\textit{et al.} and ADN but outperformed all the unpaired image denoisers. 
Even though ISCL requires the additional noise extractor for cooperative learning, its overhead (additional model size increment due to the noise extractor) is only 12.67\%. 
Besides, the number of trainable variables of ISCL used during training is only about 27\% of those of Quan~\textit{et al.} and ADN. 
We expect that the increase of the model size of ISCL may lead to even higher performance, which is left for future work.

\begin{figure}[t]
    \centering
    \includegraphics[width=0.5\textwidth]{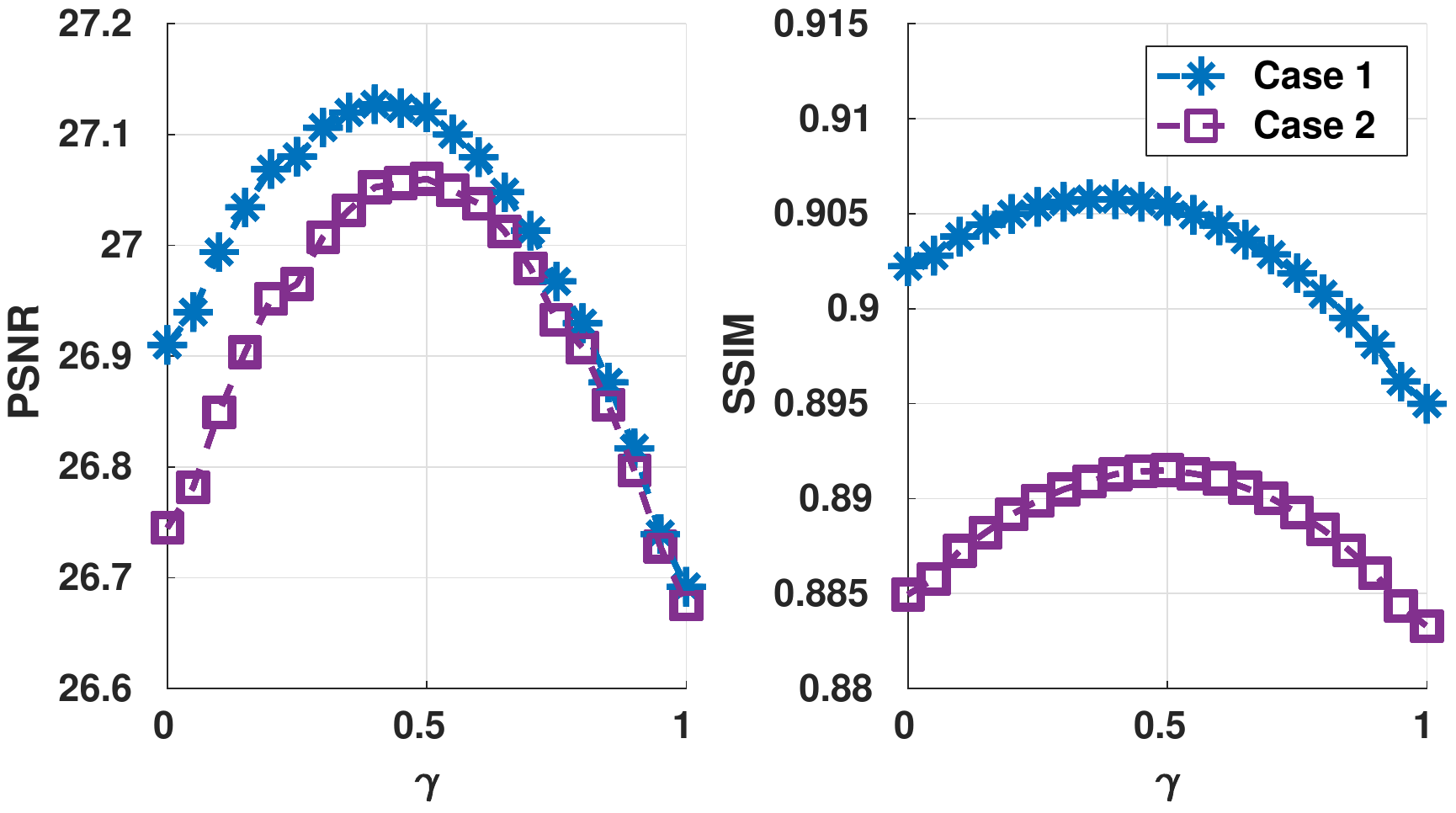}
    \caption{Performance graph in terms of $\gamma$ in EM image denoising.}
    \label{fig:fig9}
\end{figure}

We also conducted the experiment to study the effect of $\gamma$, 
which is an weighting parameter for blending between the unpaired (F) and paired (H) denoising results to generate the final ensemble result. 
As shown in Fig.~\ref{fig:fig9}, we observed that the ensemble of two clean images predicted by two different networks always achieves a better result than each single output, with the parameter value of around $\gamma=0.5$. 
%
%
We believe that the ensemble of the different network structures or different tasks reduces the variation of the error, which eventually contributes for better local optimal predictions.

\begin{figure}[t]
    \centering
    \includegraphics[width=0.48\textwidth]{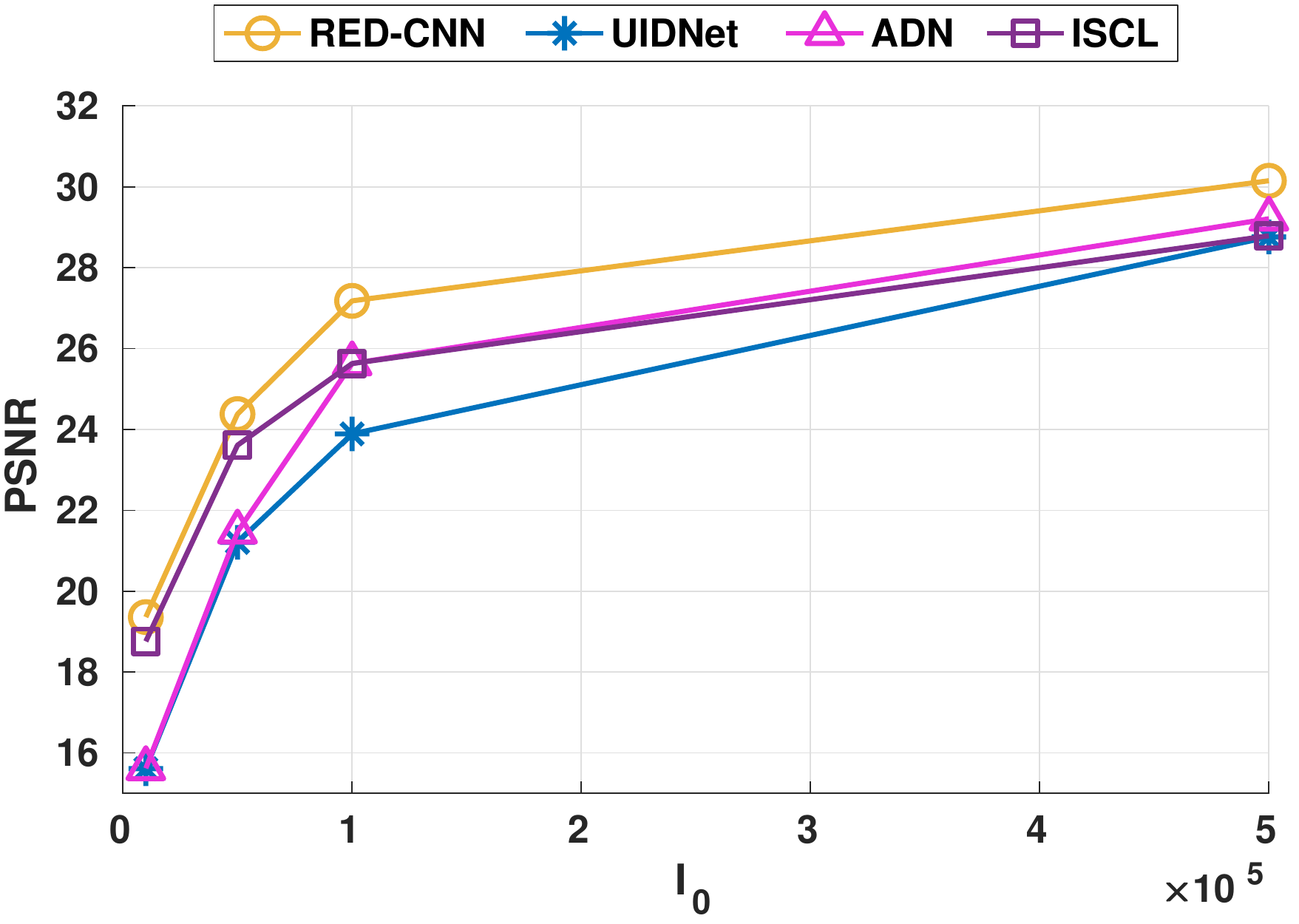}
    \caption{Performance comparison graph on various dose-levels LDCT.}
    \label{fig:fig10}
\end{figure}

Even though ISCL shows good performance in unpaired denoising, there are some limitations worth discussing.
%
%
To test the robustness to different noise levels, we conducted another experiment using different dose levels of LDCT as an extension of Section~\ref{sec:experiments}-B. 
%
For this, we retrained the models (RED-CNN, UIDNet, ADN, and ISCL) using the same training set used in the abdominal experiment of Section~\ref{sec:experiments}-B, and deployed them to LDCT images with various dose levels. 
In Table~\ref{tab:tab5}, $I_0$ denotes the unattenuated X-ray beam intensity that is usually determined by the tube electric current or voltage. In Fig.~\ref{fig:fig10}, ISCL is comparable to or outperforms other unpaired denoising methods (UIDNet and ADN) in different noise levels.
However, all the image denoising methods we tested did not recover extremely low dose CT images well (below 20db in PSNR) for $I_0=1*10^4$. 
%
The denoising approaches learned from correlation of paired or unpaired clean and noisy images have weak tolerance for unseen noise levels. 
We believe that various dose-level training samples covering the entire domain of noisy images are helpful to generalize the method and to overcome this limitation. 

Another limitation is that training the network is difficult, e.g., finding optimal hyper-parameters, due to complex loss functions and adversarial networks.
Even though we employ recent training schemes
~\cite{SWA, lookahead} to reduce the sensitivity of the hyper-parameters, training a complicated neural network as ours was not easy. 
In detail, we observed that the performance degradation occurred by excessive training iterations (due to overfitting or self-adversarial attack~\cite{advattack}) even though ours is much more robust compared to vanilla CycleGAN (see Fig.~\ref{fig:fig7}). 
We avoided such problems by employing early stopping.

\section{CONCLUSION}
In this paper, we introduced a novel denoiser, ISCL, with which the unpaired image denoising becomes feasible. 
ISCL outperformed the state-of-the-art blind denoising methods (i.e., BM3D, N2S, and N2V) and the unpaired image denoising methods (i.e., Quan \textit{et al.}, UIDNet, and ADN). 
Furthermore, ISCL showed superior performance comparable to a supervised learning-based method, which is encouraging considering ISCL is an unpaired image denoising method.
To the best of our knowledge, it is the first cooperative learning approach wherein CycleGAN and a self-residual learning-based network can complement each other under novel constraints (e.g.,  bypass-consistency, discriminator boosting, and noise-consistency). 
%
%
We discovered that the cooperative learning helps to converge faster to the optimal point than vanilla CycleGAN. 
Moreover, ISCL can arrive at better optimal point even though the network architecture is same as that of CycleGAN. 
As per our assumption in~\ref{sec:method}, we demonstrated that our proposed constraints can reduce the mapping space of prediction of CycleGAN, so that the results are closed to ground-truth. We conclude that ISCL can be applied to the real-world examples such as in the medical domain that includes complex heterogeneous noise. 
In the future, we plan to conduct clinical assessment, specifically targeting CT and MRI, to validate denoising quality and clinical applicability of ISCL. 
Extending ISCL to other image restoration applications, such as single image super-resolution, would be another interesting future research direction.


\section*{ACKNOWLEDGEMENT}
This work is supported by the Korea Health Industry Development Institute (HI18C0316), the National Research Foundation of Korea (NRF-2019M3E5D2A01063819, NRF-2021R1A6A1A13044830), the Institute for Information \& communications Technology Planning \& Evaluation (IITP-2021-0-01819), and the Korea Institute of Science and Technology (KIST) Institutional Program, Republic of Korea (2E30970).

\bibliographystyle{IEEEtran.bst}
\bibliography{reference.bib}

\begin{IEEEbiography}[{\includegraphics[width=1in,height=1.25in,clip,keepaspectratio]{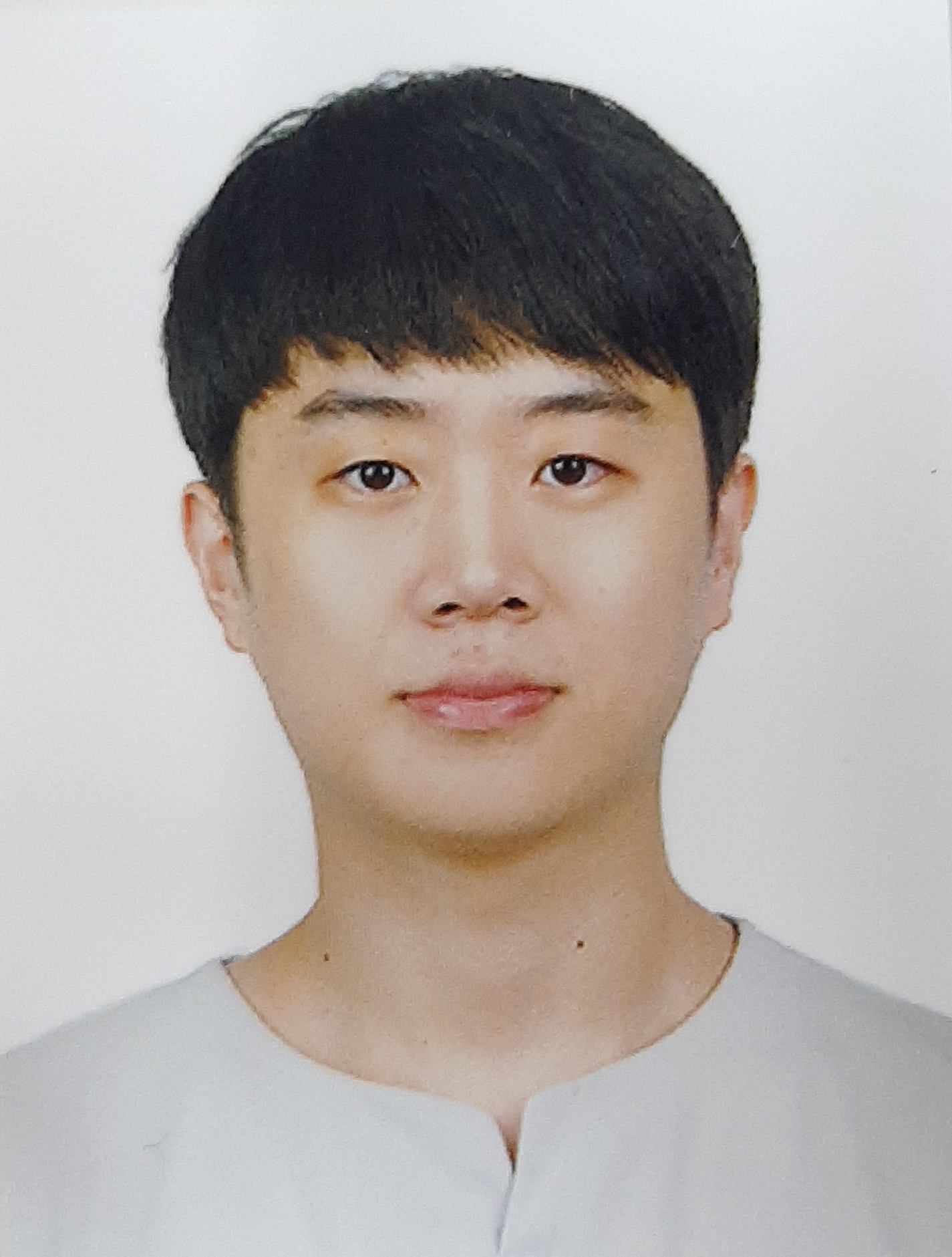}}]{Kanggeun Lee} received a B.S. degree from Ulsan National Institute of Science and Technology (UNIST), Korea in 2017. He is currently working towards a Ph.D. degree in the department of computer science and engineering at UNIST. His research interest includes image processing, machine learning, and computer vision.
\end{IEEEbiography}

\begin{IEEEbiography}[{\includegraphics[width=1in,height=1.25in,clip,keepaspectratio]{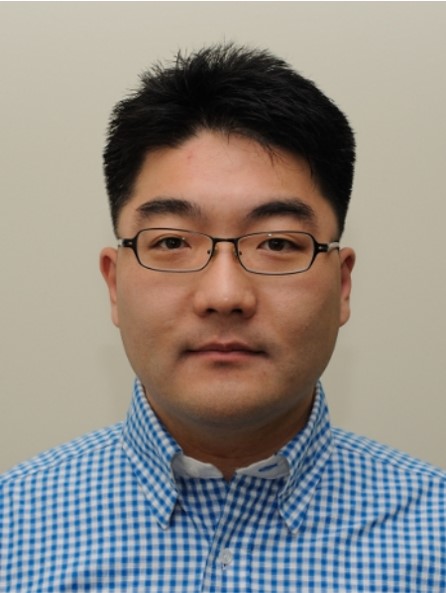}}]{Won-Ki Jeong} is currently a professor in the department of computer science and engineering at Korea University. He was an assistant and associate professor in the school of electrical and computer engineering at UNIST (2011-2020), a visiting associate professor of the neurobiology at Harvard Medical School (2017–2018), and a research scientist in the Center for Brain Science at Harvard University (2008–2011). His research interests include visualization, image processing, and parallel computing. He received a Ph.D. degree in Computer Science from the University of Utah in 2008, and was a member of the Scientific Computing and Imaging (SCI) institute.
\end{IEEEbiography}

\end{document}